\documentclass[twocolumn]{template}
\usepackage[toc,page,header]{appendix}

\usepackage{amsmath,amssymb,amsfonts}%
\usepackage{amsthm}%
\usepackage{mathrsfs}%
\usepackage{manyfoot}%
\usepackage{float}
\usepackage{adjustbox}
\usepackage{hyperref}
\hypersetup{colorlinks=true,breaklinks,linkcolor=red,citecolor=cyan}
\usepackage{minitoc}
\usepackage{algorithmic}
\usepackage[linesnumbered,ruled,vlined]{algorithm2e} 
\usepackage{listings}
\usepackage{graphicx} 
\usepackage{amssymb}
\usepackage{bbding}
\usepackage{wrapfig}
\usepackage{url}            
\usepackage{booktabs}       
\usepackage{multirow}
\usepackage{amsfonts}       
\usepackage{nicefrac}       
\usepackage{microtype}      
\usepackage[table]{xcolor}         
\usepackage{color, colortbl}
\usepackage{capt-of}
\usepackage{bm}
\usepackage{textcomp, gensymb}
\usepackage{caption}
\usepackage{enumitem, xspace}
\usepackage{makecell}
\usepackage{pifont}
\usepackage[skip=4pt]{caption}

\makeatletter
\AtBeginDocument{%
  \addtocontents{toc}{\protect\setcounter{tocdepth}{-10}}
}
\makeatother

\theoremstyle{thmstyleone}%
\theoremstyle{thmstyletwo}%

\theoremstyle{thmstylethree}%

\begin{document}

\title{Risk-Aware World Model Predictive Control for Generalizable End-to-End Autonomous Driving}


{\small
\author[1]{Jiangxin Sun}
\author[1]{Feng Xue}
\author[1]{Teng Long}
\author[1]{Chang Liu}
\author[2]{Jian-Fang Hu}
\author[2]{Wei-Shi Zheng}
\author[1]{Nicu Sebe}
}
\affiliation[1]{University of Trento, Trento, Italy}
\affiliation[2]{Sun Yat-sen University, Guangzhou, China}

\abstract{
With advances in imitation learning (IL) and large-scale driving datasets,
end-to-end autonomous driving (E2E-AD) has made great progress recently.
Currently, IL-based methods have become a mainstream paradigm:
models rely on standard driving behaviors given by experts, and learn to minimize the discrepancy between their actions and expert actions.
However, this objective of ``only driving like the expert'' suffers from limited generalization:
when encountering rare or unseen long-tail scenarios outside the distribution of expert demonstrations,
models tend to produce unsafe decisions in the absence of prior experience.
This raises a fundamental question: \textbf{\textit{Can an E2E-AD system make reliable decisions without any expert action supervision?}}
Motivated by this, we propose a unified framework named Risk-aware World Model Predictive Control (RaWMPC) to address this generalization dilemma through robust control, without reliance on expert demonstrations.
Practically, RaWMPC leverages a world model to predict the consequences of multiple candidate actions and selects low-risk actions through explicit risk evaluation.
To endow the world model with the ability to predict the outcomes of risky driving behaviors, we design a risk-aware interaction strategy that systematically exposes the world model to hazardous behaviors, making catastrophic outcomes predictable and thus avoidable.
Furthermore, to generate low-risk candidate actions at test time,
we introduce a self-evaluation distillation method to distill risk-avoidance capabilities from the well-trained world model into a generative action proposal network without any expert demonstration.
Extensive experiments show that RaWMPC outperforms state-of-the-art methods in both in-distribution and out-of-distribution scenarios, while providing superior decision interpretability.
}

\date{\today}
\correspondence{Feng Xue at \email{feng.xue@unitn.it}}
\contribution{\email{jiangxin.sun@unitn.it}}
\contribution{\email{feng.xue@unitn.it}}
\contribution{\email{teng.long@unitn.it}}
\contribution{\email{chang.liu@unitn.it}}
\contribution{\email{hujianf@mail.sysu.edu.cn}}
\contribution{\email{wszheng@ieee.org}}
\contribution{\email{nicu.sebe@unitn.it}}




\maketitle

\begin{figure*}[!t]
\centering
\includegraphics[width=1\linewidth]{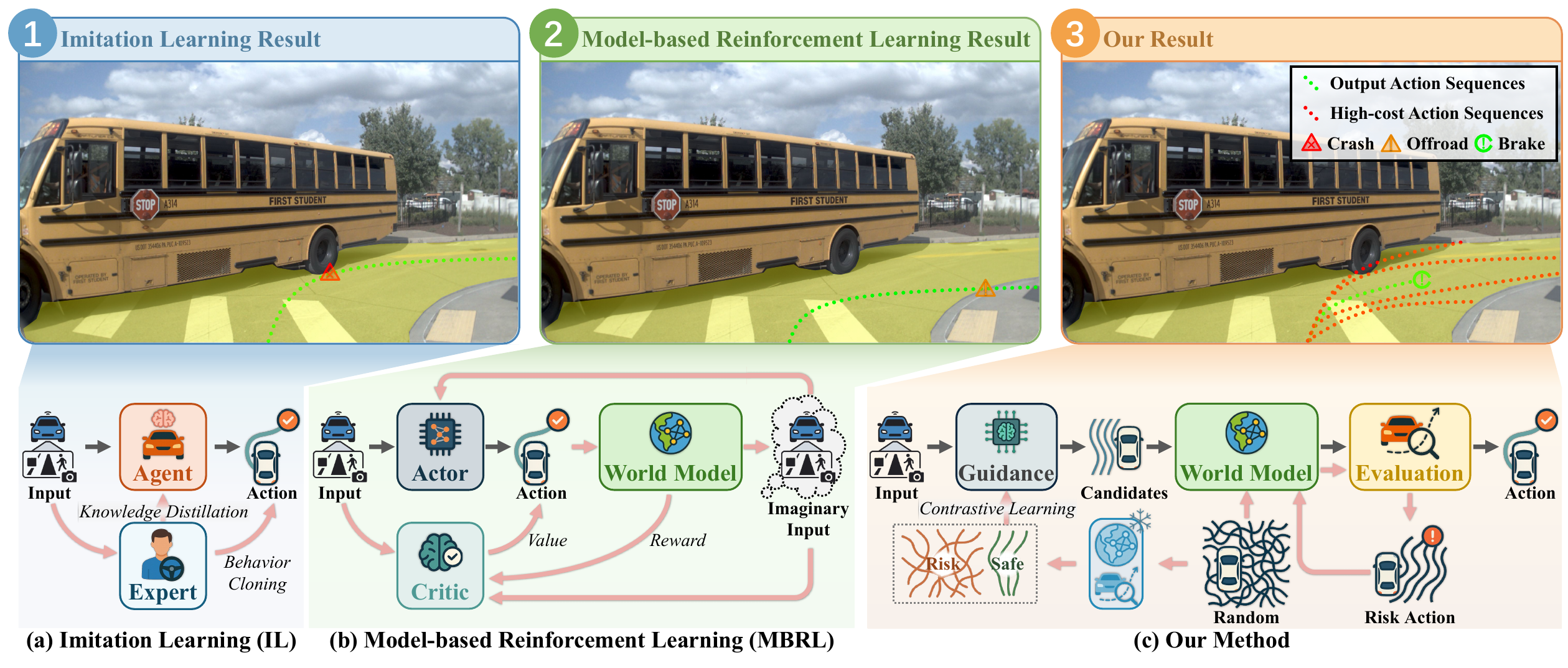}
\caption{\textbf{Comparison between existing E2E-AD methods and RaWMPC}.
The first row shows the predicted trajectories,
and the second row compares the core workflows.
Black arrows denote test-time execution,
while pink arrows indicate training-only steps.
The comparison shows that prior methods often omit explicit hazard modeling and may trigger traffic violations,
whereas RaWMPC uses a risk-aware world model to evaluate action consequences and select safe, compliant actions in critical scenes.
}
\label{fig:intro}
\end{figure*}

\section{Introduction}
\label{sec1}

End-to-end autonomous driving (E2E-AD) \cite{Reasonnet,InterFuser,TCP,TransFuser++,CarLLaVA} aims to map sensor observations to control actions,
\emph{e.g.}, steering, throttle, and brake,
without relying on hand-crafted perception, prediction, or planning modules.
Compared to traditional modular pipelines~\cite{Module1,Module2,Module3,Module4,Module5,Module6},
E2E-AD offers a more unified representation of the driving task,
enabling the policy to reason about complex interactions between the ego vehicle and dynamic environments. As such, E2E-AD has attracted growing attention due to its potential for simplified system design,
joint optimization, and real-time decision-making.

Early studies on E2E-AD \cite{MaRLn,WoR,GRIAD} primarily focused on learning driving policies via reinforcement learning (RL) through online exploration. More recent research \cite{Roach,LAV} demonstrated that \textbf{rule-based} or \textbf{RL-based agents} leveraging privileged information (\emph{e.g.}, bird’s-eye-view segmentation and high-definition maps) can produce superior driving decisions. Building upon these insights, state-of-the-art methods \cite{Reasonnet,InterFuser,TCP,ThinkTwice,CaT} generally follow an \textbf{imitation learning (IL) framework} as shown in Fig.~\ref{fig:intro}, where agents using only sensor inputs (\emph{e.g.}, RGB images and LiDAR) are trained to replicate the privileged experts’ behavior through knowledge distillation on both the driving policy and latent features.
Although some works have attempted to enhance the driving performance via future motion modeling \cite{Reasonnet,InterFuser,PlanT,Hip_ad}, action-aware future prediction \cite{MILE,ThinkTwice,LAW,WoTE}, and the
integration of large language models~\cite{Eta,VLR-Drive,Orion,Simlingo},
these approaches still adhere to the learning objective of “\textit{driving like an expert}”, as demonstrated in Fig. \ref{fig:intro} (a).
They cannot fundamentally resolve the inherent generalization dilemma of imitation learning:
\textit{Since expert demonstrations cannot cover all scenarios and situations,
imitation-based policies tend to produce unpredictable and often unsafe driving behavior when encountering unseen scenarios outside of expert demonstrations.}
More recently, \textbf{model-based RL methods}~\cite{Think2drive,Raw2Drive} have emerged and attempted to improve generalization by learning environmental dynamics and planning over them.
However, as illustrated in Fig. \ref{fig:intro} (b),
most of them still aim to maximize the expected reward and lack explicit modeling and sampling of rare but high-risk situations,
and thus continue to struggle to guarantee safety in these scenarios.

In this paper,
we argue that \emph{``enabling an E2E-AD system to learn and proactively avoid risky actions is more important than replicating expert driving behavior verbatim''}.
Motivated by this perspective,
we propose an E2E-AD framework that does not require any expert action supervision,
called \textbf{Risk-aware World Model Predictive Control} (RaWMPC), as shown in Fig. \ref{fig:intro}(c).
This framework discards expert demonstrations and instead leverages a risk-aware world model to drive predictive control to overcome the generalization challenge.
Different from model-based RL that trains an actor to maximize reward from imagined rollouts of a world model,
the world model in RaWMPC predicts near-future states for a set of ``candidate'' driving behaviors and explicitly evaluates their risk, so as to select the lowest-risk candidate.
To endow our world model with risk-awareness,
we introduce a \textbf{risk-aware interaction} strategy:
Starting from scratch, the model selects self-identified high-risk actions to interact with the environment,
from which our world model learns to predict the consequences of diverse risky behaviors.
Without any expert demonstration, our world model can reach strong performance from scratch,
and can be further accelerated if a few video clips are provided by benchmarks for a light warm-up.
Finally, to efficiently provide low-risk candidates at test time, we further propose a \textbf{self-evaluation distillation} for driving policy learning.
The well-trained world model is leveraged to identify safe and risky behaviors from the sampled action space, and to distill this knowledge, via safety–risk contrastive learning, into a generative action proposal network.
Experiments on Bench2Drive and NAVSIM show that RaWMPC, even without expert demonstrations, surpasses previous state-of-the-art methods, while the optional light warm-up further helps to accelerate the convergence and improve performance.
More importantly,
since RaWMPC learns risk-awareness from interaction rather than expert labels,
it achieves substantially higher driving performance in previously unseen scenarios.
Our main contributions can be summarized as follows:
\begin{itemize}
\item We propose \textbf{RaWMPC}, an E2E-AD framework with zero expert requirement.
Unlike IL and MBRL methods, RaWMPC uses a world model to select low-risk behaviors from multiple candidate actions, which naturally improves the interpretability and reliability of its decisions.
\item Within RaWMPC, we design a \textbf{risk-aware interaction} learning strategy that enables the world model to acquire risk-awareness purely from environment interaction, without any expert demonstration.
\item We introduce a \textbf{self-evaluation distillation} scheme for driving policy learning, which provides high-quality candidate actions at test time, even outperforming policies learned directly from expert demonstrations.
\end{itemize}

\section{Related Work}

\subsection{End-to-End Learning in Autonomous Driving}

Learning-based autonomous driving approaches typically follow two main paradigms: \textbf{imitation learning (IL)} and \textbf{reinforcement learning (RL)} \cite{Survey1,Survey2}. 
Early studies explored RL extensively \cite{MaRLn,WoR,GRIAD,RL1,RL2,RL3,SPC,IPC,RL5,RL6} for its natural capability to refine driving strategies through interactive feedback. 
Subsequent works \cite{Roach,Think2drive} demonstrated that training sensor-based agents to imitate RL experts' behavior could lead to superior performance, prompting a shift toward leveraging privileged information (e.g., BEV segmentation and HD maps) to train stronger RL experts.
More recently, model-based RL has revisited this line by learning explicit dynamics/world models and performing look-ahead rollouts \cite{Raw2Drive}, improving consequence-aware evaluation and sample efficiency. 
Nevertheless, these RL approaches are commonly driven by maximizing expected return, and they rarely provide an explicit mechanism to systematically discover and model rare-but-catastrophic outcomes, making reliable decisions in long-tail, high-risk scenarios still challenging.

With large-scale driving datasets, imitation learning has attracted increased attention and achieved state-of-the-art performance in closed-loop autonomous driving \cite{Simlingo,Goalflow,Hip_ad,HydraNeXt}. Most IL approaches rely on collecting trajectory data and features from RL-based \cite{Roach,Think2drive} or rule-based \cite{Transfuser,Drivelm} experts using privileged information, and train sensor-based IL agents to replicate expert behaviors through knowledge distillation. 
A variety of research studies have explored ways to improve IL performance, such as multi-modal information fusion \cite{Transfuser,TransFuser++,Transfuser_conf}, object motion modeling \cite{Reasonnet,PlanT,InterFuser,Drivetransformer}, action-aware future prediction \cite{MILE,ThinkTwice,LAW,WoTE}, feature alignment \cite{Driveadapter,CaT}, and the integration of large language models \cite{Drivelm,Lmdrive,Drivegpt4,Rag,CarLLaVA,Lingoqa,Simlingo,Orion}.
Despite impressive results, the core objective of ``driving like the expert'' inherently limits generalization: expert demonstrations cannot cover all long-tail situations, and experts typically avoid dangerous behaviors, leaving IL agents with limited supervision on how to recognize and proactively avoid high-risk actions. 
Moreover, pure imitation often provides limited interpretability because it outputs a single action without explicitly comparing alternative actions via consequence evaluation. 
Motivated by these insights, we propose RaWMPC, a unified framework that replaces expert action supervision with risk-aware predictive control: it learns a risk-aware world model via a risk-aware interaction strategy that deliberately exposes the model to risky behaviors, and uses the learned model to predict and evaluate the consequences of multiple candidate actions, selecting low-risk behaviors with explicit risk evaluation. 

\subsection{World Models}

World models approximate environment transitions under the Markov Decision Process and have demonstrated success in RL \cite{RLWM1,RLWM2,RLWM4,RLWM5,RLWM6,RLWM7,RLWM9,RLWM10,MaRLn} by predicting future states and rewards from current observations and actions.
However, applying these models to complex tasks such as autonomous driving remains challenging.
Previous research \cite{Gaia1,Vista,WM1,WM2,WM3,WM4,WM5,WM6,WM7,WM8,WM9,WM10,DriveDreamer4D,ReconDreamer,MaskGWM,GaussianWorld} has mainly leveraged world models to generate controllable future driving trajectories (e.g., RGB images and 3D/4D representations) conditioned on specific actions and scene descriptions. 
Such predictive models can enlarge training data and increase diversity, potentially benefiting downstream imitation learning, especially for rare scenes such as traffic accidents.

Beyond prediction, a few recent attempts have utilized world models to improve closed-loop autonomous driving performance. In particular, some works have started to connect world modeling with planning and online evaluation in driving settings \cite{DrivingGPT,Epona,WoTE},
while others provide high-fidelity generative platforms that enable closed-loop evaluation \cite{Navsim,DriveArena,Navsim_v2}. 
In particular, model-based IL methods \cite{MILE,LAW,WoTE,Drivedpo} employ world models to support better imitation: LAW \cite{LAW} enhances end-to-end driving by predicting future information in a latent world model to assist policy learning, and WoTE \cite{WoTE} performs online trajectory evaluation via a BEV world model to score candidate trajectories. 
More recently, model-based RL methods have drawn increasing attention. Think2Drive \cite{Think2drive} designed a model-based RL expert to forecast action-conditioned future rewards for training a more effective critic network, and Raw2Drive \cite{Raw2Drive} leveraged a world model pretrained from privileged experts to guide the learning of sensor agents. 
Despite these advances, most existing works still inherit supervision from experts or rewards, and they largely focus on imitation fidelity or expected-return maximization, lacking explicit mechanisms to systematically discover, model, and avoid rare-but-high-risk outcomes. In contrast, RaWMPC uses the world model as a risk evaluator within predictive control: we introduce a risk-aware interaction strategy to intentionally explore risky behaviors so that catastrophic consequences become predictable and avoidable, and we select actions by explicitly minimizing risk over multiple candidates, enhancing the interpretability, reliability, and generalization of the decision-making process.

\begin{figure*}[t]
    \centering
    \includegraphics[width=1\linewidth]{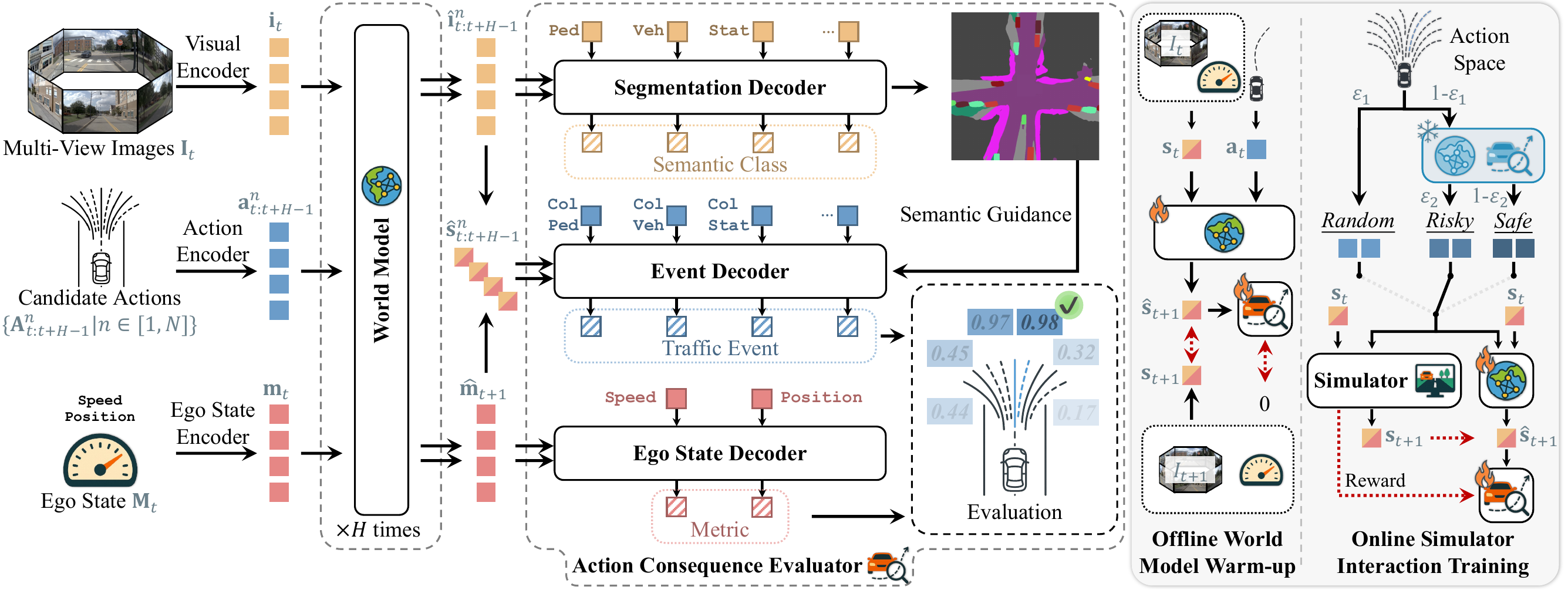}
    \caption{
\textbf{Overview of RaWMPC.}
Multi-view images $\mathbf{I}_t$, ego state $\mathbf{M}_t$,
and candidate action sequences $\{\mathbf{A}^{n}_{t:t+H-1}\}_{n=1}^{N}$ are encoded and rolled out by a world model over horizon $H$.
Three decoders predict semantic segmentation,
semantic-guided traffic events,
and future ego states, enabling action evaluation for predictive control.
Training combines offline warm-up on logged trajectories with online simulator interaction using world-model-guided exploration.
}
    \label{fig:framework}
\end{figure*}

\section{Method}
To demonstrate our solution, Sec. \ref{sec:RaWMPC} introduces the overall network structure and pipeline of our RaWMPC. Sec.~\ref{sec:RaIT} presents our training scheme, \emph{i.e.}, risk-aware interactive training, for efficient optimization. To ensure the whole E2E-AD system runs efficiently during testing, Sec. \ref{sec:guidance} illustrates a self-evaluation distillation method to train an action proposal network.

\subsection{Network Structure of RaWMPC}
\label{sec:RaWMPC}

\subsubsection{Problem Setup}
\label{sec:pipeline}
We consider a closed-loop end-to-end autonomous driving (E2E-AD) setting.
At each time step $t$, the agent receives three inputs:
a visual input $\mathbf{I}_t$ (\textit{multi-view RGB images}),
an ego-centric measurement $\mathbf{M}_t$ (\textit{velocity and position}), and
a set of candidate driving behaviors $\{\mathbf{A}^{n}_{t:t+H-1}\}_{n=1}^{N}$,
where $N$ is the number of candidates and $H$ is the planning horizon.
Each step of action consists of three values:
$\mathbf{A}=(\texttt{steer}\in[-1,1],\texttt{throttle}\in[0,1],\texttt{brake}\in[0,1])$.
Based on the driving history $(\mathbf{I}_{1:t},\mathbf{M}_{1:t},\mathbf{A}_{1:t-1})$,
RaWMPC aims to select the best one $\mathbf{A}^{n^\star}_{t:t+H-1}$ from the candidates, so that the vehicle can move toward a destination while ensuring safety and compliance with traffic rules.

\subsubsection{Overview}
As illustrated in Fig. \ref{fig:framework},
RaWMPC begins from input encoding.
A visual encoder, an action encoder and an ego-state encoder are used to map $\mathbf{I}_t$, $\{\mathbf{A}^{n}_{t:t+H-1}\}_{n=1}^{N}$,
and $\mathbf{M}_t$ into embeddings $\mathbf{i}_t$, $\{\mathbf{a}^{n}_{t:t+H-1}\}_{n=1}^{N}$, and $\mathbf{m}_t$, respectively.
Then, based on the observed states $\mathbf{s}_{1:t}\!=\!(\mathbf{i}_{1:t},\mathbf{m}_{1:t}\!)$,
we use a world model to estimate its future state $\hat{\mathbf{s}}^{n}_{t+1:t+H}$ conditioned on each action embeddings $\mathbf{a}^{n}_{t:t+H-1}$.
Finally, we select the action that enables the ego vehicle to advance safely while avoiding traffic infractions, by decoding $\hat{\mathbf{s}}^{n}_{t+1:t+H}$ and computing a cost value:
\begin{equation}
\begin{aligned}
\label{eq:action_selection}
&\mathbf{A}^{\star}_{t:t+H-1} = \mathbf{A}^{n^\star}_{t:t+H-1}, \\
&\text{where}\,\,\, n^\star = \underset{n\in\{1,\dots,N\}}{\arg\min} C\!\big(\hat{\mathbf{s}}^{n}_{t+1:t+H}\big).
\end{aligned}
\end{equation}
where $C(\cdot)$ denotes the cost function in the decoding process (will be detailed in Eq.~\eqref{eq_cost}),
and $n^\star$ is the index of optimal action.
Compared to imitation learning schemes,
RaWMPC offers improved interpretability and introduces an explicit mechanism for decision validation and risk mitigation.

\subsubsection{World Model}
In our pipeline,
the world model, denoted as $\mathcal{M}$, is employed to predict future states given an action.
Specifically,
conditioned on the observed states $\mathbf{s}_{1:t}\!=\!(\mathbf{i}_{1:t},\mathbf{m}_{1:t}\!)$ and the potential next action $\mathbf{a}^{n}_{t}$,
the world model $\mathcal{M}$ predicts a near-future state $\hat{\mathbf{s}}^{n}_{t+1}$.
Then, for the further-future time step $\{2,\dots,H\}$,
the world model recursively rolls out, which can be formalized as an autoregressive factorization:
\begin{equation}
\begin{aligned}
\setlength{\abovedisplayskip}{2pt}
&p_\mathcal{M}(\hat{\mathbf{s}}^{n}_{t+1:t+H}|\mathbf{s}_{1:t},\mathbf{a}^{n}_{1:t+H-1}) =\\
&\prod\nolimits_{k=1}^{H}
p_\mathcal{M}\Big(\hat{\mathbf{s}}^{n}_{t+k}| (\mathbf{s}_{1:t}, \hat{\mathbf{s}}^{n}_{t+1:t+k-1})\,\,,\, \mathbf{a}^{n}_{1:t+k-1}\Big)
\setlength{\abovedisplayskip}{2pt} 
\end{aligned}
\end{equation}
where $p_\mathcal{M}$ denotes the conditional distribution defined by the world model $\mathcal{M}$.
$\mathbf{a}^{n}_{1:t+k-1}=[\mathbf{a}_{1:t-1},\mathbf{a}^{n}_{t:t+k-1}]$ denotes a candidate action sequence containing action history.

\subsubsection{Semantic-Guided Decoding}
Once the world model predicts a sequence of future states, $\hat{\mathbf{s}}^{n}_{t+1:t+H}$,
three transformer decoders separately map these states to task-specific outputs:
semantic segmentation, potential traffic events (\emph{e.g.}, collision), and future ego-states (\emph{e.g.}, position).
In what follows, we describe these decoders using the predicted state at time step $t+k$, \emph{i.e.}, $\hat{\mathbf{s}}^{n}_{t+k}$, where $k \in \{1,\dots,H\}$.

To enable a higher-level understanding of driving scenes and provide visual explanations for predicted traffic events,
we inject semantic attention from the segmentation decoder into the event decoder.
The \textbf{segmentation} decoder adopts a standard transformer attention:
\begin{equation}
\!\!\mathtt{Att_{seg}}(\!\mathbf{Q}_c,\!\mathbf{K}_c,\!\!\mathbf{V}_c\!) \!=\! \mathtt{softmax}(\mathtt{sim}(\mathbf{Q}_c,\mathbf{K}_c))\!\cdot\!\!\mathbf{V}_c,
\end{equation}
where $\mathbf{Q}_c$ are learnable class queries, and $\mathbf{K}_c,\mathbf{V}_c$ are derived from visual tokens $\hat{\mathbf{i}}^{n}_{t+k}\subset \hat{\mathbf{s}}^{n}_{t+k}$.
In the final layer, we follow SegViT~\cite{SegViT} to predict the one-hot semantic segmentation map $\hat{\mathbf{Y}}^{n}_{t+k}$ for each input class query.
We then augment the \textbf{event} decoder by fusing its attention logits with the corresponding semantic attention logits from the last segmentation layer:
\begin{equation}
\begin{aligned}
\mathbf{Z}_e &= \mathbf{Q}_e \mathbf{K}_e^\top, \,\,\,\mathbf{Z}_c = \mathrm{pad}(\mathbf{Q}_c \mathbf{K}_c^\top), \\
\hat{E}^{n}_{t+k}
&= \mathrm{sigmoid}\!\left(
    \mathrm{softmax}\!\left(
        \mathbf{W}_e * [\mathbf{Z}_e, \mathbf{Z}_c]
    \right) \!\mathbf{V}_e
   \right).
\end{aligned}    
\end{equation}
where $\mathbf{Q}_e$ are learnable event queries,
and $\mathbf{K}_e,\mathbf{V}_e$ are computed from predicted future states $\hat{\mathbf{s}}^{n}_{t+k}$.
$\mathtt{pad}(\cdot)$ zero-pads $\mathbf{Z}_c$ to match the size of $\mathbf{Z}_e$,
and $\mathbf{W}_e$ is a $1\times1$ convolution that fuses the concatenated logits.
The output of the event decoder, $\hat{E}^{n}_{t+k}\in[0,1]^\alpha$, represents the probabilities of $\alpha$ event types.
Finally, for the future \textbf{ego-state} prediction, we decode the ego token $\hat{\mathbf{m}}^{n}_{t+k}\subset \hat{\mathbf{s}}^{n}_{t+k}$ to obtain the speed and position $\hat{\mathbf{M}}^{n}_{t+k}$.

Guided by the semantic attention map, the event decoder draws more attention to regions critical to specific events.
For instance, when recognizing the vehicle collision event,
the model focuses more on vehicle areas, improving the accuracy and reliability of event predictions.

\subsubsection{Action Selection and Predictive Control}
Given the decoder outputs, we perform \emph{predictive control} by evaluating each candidate action sequence over the planning horizon $H$ and selecting the one with the minimum predicted cost.

Specifically, for the $n$-th candidate $\mathbf{A}^n_{t:t+H-1}$,
we consider (i) progress toward the target and (ii) the risk of traffic-violation events.
Let $\mathbf{p}^\star$ be the target 3D position and $\hat{\mathbf{p}}^{n}_{t+k}\subset\hat{\mathbf{M}}^{n}_{t+k}$ the predicted ego position at step $t+k$.
We define the progress as the reduction in target distance:
\begin{equation}
\hat{D}^{n}_{t+k} = \big\| \mathbf{p}^\star - \hat{\mathbf{p}}^{n}_{{t+k}-1} \big\|_2 - \big\| \mathbf{p}^\star - \hat{\mathbf{p}}^{n}_{{t+k}} \big\|_2,
\end{equation}
We then define the predictive-control objective as:
\begin{equation}
\setlength{\abovedisplayskip}{2pt}
\label{eq_cost}
\begin{aligned}
\!\!\!C(\hat{\mathbf{s}}^{n}_{t+1:t+H}) = \!\!\sum_{k=1}^H \eta_k (-\hat{D}^{n}_{t+k} + \!\!\sum_{j=1}^{\alpha} \lambda_j\, \hat{E}^{n}_{t+k,j}), \\
\end{aligned}
\setlength{\belowdisplayskip}{2pt}
\end{equation}
where $\eta_k=\max(2^{-k+1},1/8)$ down-weights distant predictions to account for increasing uncertainty. We floor $\eta_k$ at $1/8$ to avoid vanishing contributions from distant steps, which stabilizes planning when $H$ is moderately large. $\lambda_j>0$ reflects the severity of violation type $j$ (\textit{e.g.}, pedestrian/vehicle collisions receive larger weights).
Finally, we select the action sequence that minimizes the horizon cost in Eq.~\eqref{eq_cost},
\textit{i.e.}, a model-predictive control policy that favors faster progress while proactively reducing the probability of predicted violations.

\subsubsection{Overall Loss of RaWMPC}
The RaWMPC framework is trained in an end-to-end manner with a world model loss $\mathcal{L}_{\text{world}}$, a segmentation loss $\mathcal{L}_{\text{seg}}$ from SegViT \cite{SegViT}, ego-state loss $\mathcal{L}_{\text{ego}}$, and event loss $\mathcal{L}_{\text{event}}$:
\begin{equation}
\setlength{\abovedisplayskip}{3pt}
\mathcal{L} = \mathcal{L}_{\text{world}} + \mathcal{L}_{\text{seg}} + \mathcal{L}_{\text{ego}} + \mathcal{L}_{\text{event}}.
\setlength{\belowdisplayskip}{3pt}
\end{equation}
Following SegViT~\cite{SegViT}, the segmentation term includes classification loss $\mathcal{L}_{\text{cls}}$ (cross-entropy) and the binary mask loss. The mask loss consists of a focal loss $\mathcal{L}_{\text{focal}}$~\cite{Focal} and a dice loss $\mathcal{L}_{\text{dice}}$~\cite{Dice} for optimizing the segmentation accuracy:
\begin{equation}
\setlength{\abovedisplayskip}{2pt}
\label{eq:seg}
\mathcal{L}_{\text{seg}}=\mathbf{\mathcal{L}}_{\text{cls}}+\mathbf{\mathcal{L}}_{\text{focal}}+\mathbf{\mathcal{L}}_{\text{dice}}.
\setlength{\belowdisplayskip}{2pt}
\end{equation}
For the other three losses,
we use Mean Squared Error (MSE) to supervise the ego-state and world model supervision,
and take Binary Cross Entropy (BCE) loss for the event decoder:
\begin{align}
\setlength{\abovedisplayskip}{2pt}
\label{eq:meas}
\mathcal{L}_{\text{world}}&=\frac{1}{H}\sum\nolimits_{k=1}^{H}\left\| \hat{\mathbf{s}}_{t+k}-\mathbf{s}_{t+k}\right\|_2^2, \nonumber \\
\mathcal{L}_{\text{ego}}&=\frac{1}{H}\sum\nolimits_{k=1}^{H}\left\| \hat{\mathbf{M}}_{t+k}-\mathbf{M}_{t+k}\right\|_2^2, \\
\mathcal{L}_{\text{event}}&=\frac{1}{H}\sum\nolimits_{k=1}^{H} BCE(\hat{E}_{t+k}, \mathbf{E}_{t+k}), \nonumber
\setlength{\belowdisplayskip}{2pt}
\end{align}
where $BCE(\cdot)$ denotes the BCE loss.
All the annotations $\mathbf{s}_{t+k}, \mathbf{M}_{t+k}, \mathbf{E}_{t+k}$ along the executed rollout under $\mathbf{A}_{t:t+H-1}$ are obtained from the simulator (\textit{e.g.}, CARLA).

\subsection{Risk-aware Interactive Training}
\label{sec:RaIT}
To enable RaWMPC to evaluate diverse actions and identify risky scenarios,
we propose a two-stage \emph{risk-aware} interactive training scheme, as shown in Fig. \ref{fig:framework}.
We first warm-start the world model from logged driving trajectories.
Then, we refine it via online simulator interaction,
intentionally collecting both \textbf{good} (safe, goal-directed) and \textbf{bad} (hazardous) rollouts to improve generalization under out-of-distribution controls and to learn rare but safety-critical events. 
Notably, RaWMPC does not rely on expert action labels for policy learning.
The optional warm-up stage, when used, serves solely to initialize the predictive world model from observed state transitions, rather than to imitate expert actions.


\subsubsection{Offline World Model Warm-up}
We bootstrap RaWMPC using a small set of logged trajectories to achieve simple and basic state-forecasting capability.
Given state-action sequences $\{(\mathbf{s}_t,\mathbf{a}_t)\}_{t=1}^{T}$ from NAVSIM or CARLA,
the world model predicts the next state $\hat{\mathbf{s}}_{t+1}$ and is trained with $\mathcal{L}_{\text{world}}$ to match the ground-truth $\mathbf{s}_{t+1}$.
We supervise the segmentation and ego-state decoders using the simulator-provided annotations.
Since the warm-up trajectories contain no traffic violations,
we train the event decoder with an all-zero target.
In this way, only a small subset of training data (10\%) is typically sufficient for warm-up, providing a reliable initialization for long-horizon rollouts and stabilizing subsequent world model optimization.

\begin{figure}
\centering
\includegraphics[width=\linewidth]{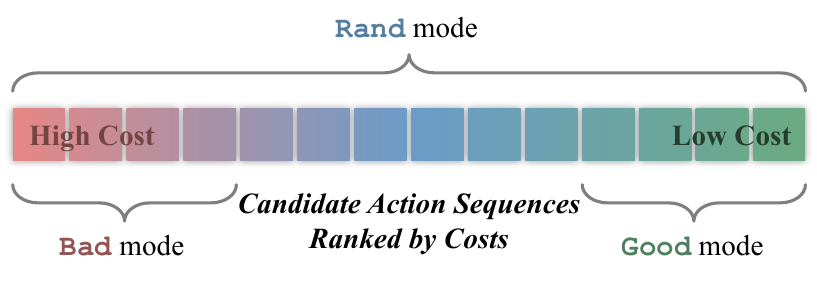}
\caption{\textbf{Different action-selection ranges under three driving modes in online simulator interaction.}
Red denotes high cost and green denotes low cost.
\texttt{rand} samples uniformly from all candidates,
\texttt{bad} samples from the high-cost region,
and \texttt{good} samples from the low-cost one.
}
\label{fig:interactive}
\end{figure}

\subsubsection{Online Simulator Interactive Training}
Offline warm-up data are mostly concentrated around human-like safe behaviors and thus provide limited coverage of hazardous or unconventional actions.
To learn the consequences of risky behaviors, we perform \emph{world-model-guided exploration}:
selected simulator rollouts are fed back to refine the same world model,
progressively improving prediction fidelity and risk sensitivity.

Specifically,
to ensure temporal continuity and avoid unrealistic control jitter,
we sample horizon-$H$ action sequences (segment-wise) rather than single-step actions (step-wise).
The segment-wise sampling allows sustained safe or risky behaviors to unfold and reveals their long-term consequences.
At each training step,
we sample $N_s$ horizon-$H$ candidate action sequences $\{\mathbf{A}^n_{t:t+H-1}\}_{n=1}^{N_s}$,
roll out future states $\hat{\mathbf{s}}^{n}_{t+1:t+H}$ with the current world model,
evaluate their costs $\{C^n\}$ using Eq. \eqref{eq_cost}, and rank candidates by costs.

\noindent
\textbf{Modes for Interaction.} We define three patterns for our risk-aware sampling strategy to select one candidate from $\{\mathbf{A}^n_{t:t+H-1}\}_{n=1}^{N_s}$ to execute, as shown in Fig. \ref{fig:interactive}:
\begin{itemize}
\item \texttt{Rand} samples uniformly from all candidates;
\item \texttt{Bad} samples from high-cost candidates;
\item \texttt{Good} samples from low-cost candidates.
\end{itemize}
At the start of segment $r$,
the practical control mode is sampled according to probability:
\begin{equation}
m_r =
\begin{cases}
\texttt{rand}, & \text{w.p. } \varepsilon_1,\\
\texttt{bad},  & \text{w.p. } (1-\varepsilon_1)\varepsilon_2,\\
\texttt{good}, & \text{w.p. } (1-\varepsilon_1)(1-\varepsilon_2),
\end{cases}
\end{equation}
where ``w.p.'' means ``with probability''. $\varepsilon_1$ controls broad action-space exploration, and $\varepsilon_2$ controls the fraction of risk-seeking interaction within model-guided sampling.

\noindent
\textbf{Soft Candidates Selection in Three Modes.}
Given sorted candidate actions with costs $\{C^n\}$, we construct two cost-quantile sets: $\mathcal{N}_{\texttt{good}}$ as the bottom-$K$ candidates and $\mathcal{N}_{\texttt{bad}}$ as the top-$K$ candidates.
To avoid low-information trajectories,
we filter $\mathcal{N}_{\texttt{bad}}$ by removing degenerate rollouts (\textit{e.g.}, those caused by unrealistic excessive control jumps),
yielding $\tilde{\mathcal{N}}_{\texttt{bad}}$. In the \texttt{rand} mode, we randomly select the executing action sequence from all candidates.

In the \texttt{good} mode, rather than \textbf{deterministically} selecting the minimum-cost candidate, we sample from $\mathcal{N}_{\texttt{good}}$ using a \textbf{soft distribution} to preserve diversity among low-cost plans and mitigate bias from imperfect model predictions:
\begin{equation}
P(n \mid \texttt{good}) \propto \exp(-C^n/\tau_g), \quad n \in \mathcal{N}_{\texttt{good}}.
\end{equation}
where $\tau_g$ is a temperature hyper-parameter that controls the softness of the sampling distributions in the \texttt{good} mode, trading off greediness for diversity.
This stochastic selection avoids repeatedly executing a single estimated optimum and encourages broader coverage of nominal behaviors.

In the \texttt{bad} mode,
instead of always executing the maximum-cost trajectory,
we sample from high-cost candidates to deliberately expose the model to a spectrum of risky outcomes that are under-represented in safe logs:
\begin{equation}
P(n \mid \texttt{bad}) \propto \exp(C^n/\tau_b), \quad n \in \tilde{\mathcal{N}}_{\texttt{bad}}.
\end{equation}
where $\tau_b$ is a temperature hyper-parameter.
Compared to argmax selection, this soft sampling strategy prevents over-concentration on extreme or degenerate failures while still biasing interaction toward high-risk regions.

In this way, segment-wise interaction and soft cost-based sampling bias exploration toward temporally coherent and informative safe and hazardous trajectories, enabling the world model to learn both reasonable dynamics and safety-critical consequences for risk-aware decision making.

\begin{figure}
\centering
\includegraphics[width=\linewidth]{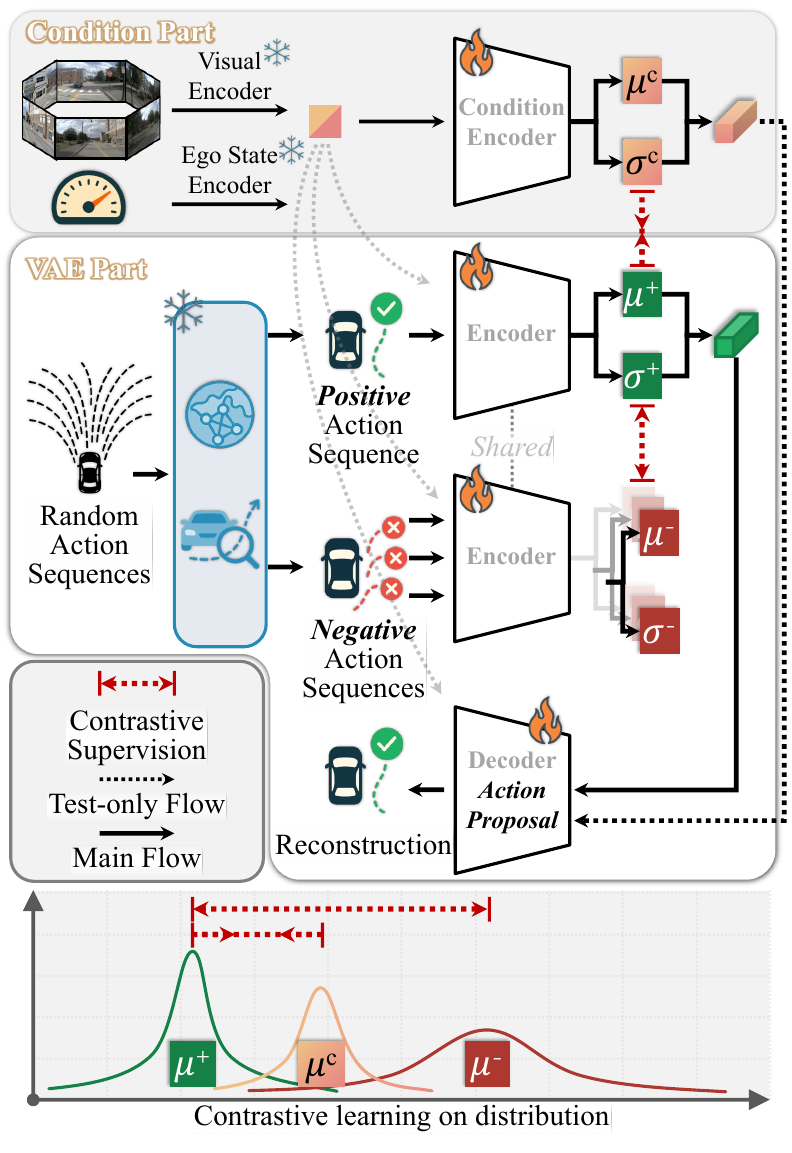}
\caption{\textbf{Self-Evaluation Distillation for Policy Learning.}
A cVAE is trained with RaWMPC-scored actions in a contrastive manner,
pulling the condition prior toward positives and pushing it away from negatives.
The well-trained decoder serves as the test-time action proposer.}
\label{fig:guidance}
\end{figure}

\subsection{Self-Evaluation Distillation for Policy Learning}
\label{sec:guidance}
After risk-aware interactive training,
RaWMPC can reliably \emph{score} candidate action sequences by predicting their long-horizon consequences.
To reduce the cost of online optimization at test time, we distill this evaluation capability into a lightweight \emph{action proposal} network, enabling efficient inference \emph{without expert demonstrations}.
The action proposal network corresponds to the ``Guidance'' module illustrated in Fig.~\ref{fig:intro}(c), and is used to generate candidate action sequences for predictive control.
Our key idea is to use RaWMPC as a self-evaluator to pseudo-label sampled actions and train a generative policy via contrastive learning.

\subsubsection{Action Sampling and Pseudo-labeling}
Given a state history $\mathbf{s}_{1:t}$ (simplified as $\mathbf{s}$),
we randomly sample $N_s$ horizon-$H$ action sequences $\{\mathbf{A}^n_{t:t+H-1}\}_{n=1}^{N_s}$ (simplified as $\{\mathbf{A}^n\}$) and compute their costs $\{C^n\}$ with the pretrained RaWMPC.
We then form pseudo labels by ranking costs: the lowest-cost sequence is treated as a positive example $\mathbf{A}^{+}$, and the top-$K$ highest-cost sequences are treated as negatives $\{\mathbf{A}^{-}_j\}_{j=1}^{K}$.
This construction transfers RaWMPC's knowledge (low-risk / high-quality actions) to the proposal network while avoiding any external supervision.

\subsubsection{Action Proposal Network}
Following \cite{Trajectron++,Agentformer},
we adopt a conditional VAE (cVAE) with an action encoder $q_\theta(z|\mathbf{A},\mathbf{s})$,
a conditional prior $p_\gamma(z|\mathbf{s})$,
and a decoder $p_\psi(\mathbf{A}|z,\mathbf{s})$.
The decoder serves as the proposal policy at inference.

For the positive action, we obtain a Gaussian posterior
$q^{+}=q_\theta(z|\mathbf{A}^{+},\mathbf{s})=\mathcal{N}(\mu^{+},\mathrm{diag}((\sigma^{+})^2))$ and train the decoder to reconstruct $\mathbf{A}^{+}$.
For each negative action $\mathbf{A}^{-}_j$, we compute
$q^{-}_j=q_\theta(z|\mathbf{A}^{-}_j,\mathbf{s})=\mathcal{N}(\mu^{-}_j,\mathrm{diag}((\sigma^{-}_j)^2))$, but \emph{\textbf{do not}} reconstruct negatives to prevent the generator from imitating unsafe behaviors.
The conditional prior is $p^{c}=p_\gamma(z|\mathbf{s})=\mathcal{N}(\mu^{c},\mathrm{diag}((\sigma^{c})^2))$.

\subsubsection{Contrastive Training Objective}
To address the lack of expert supervision in policy learning,
we use an InfoNCE objective to make the conditional prior predictive of high-quality actions.
Concretely, there are two potential contrastive formulations:
\begin{itemize}
    \item Using $p^{c}$ as the anchor, pulling $p^{c}$ toward $q^{+}$ while pushing it away from $\{q^{-}\}_{j=1}^{K}$.
    \item Using $q^{+}$ as the anchor, pulling $q^{+}$ toward $p^{c}$ while pushing it away from $\{q^{-}\}_{j=1}^{K}$.
\end{itemize}
We empirically found that the former often produces under-optimized trajectories.
One possible reason is that negative samples are far more numerous and broadly cover the latent space,
therefore $p^{c}$ is easily driven to a region that is far from most negatives yet not sufficiently close to the positive.
In contrast, the latter explicitly pulls $p^{c}$ toward the statistical center of the positive posterior and,
via $q^{+}$, indirectly separates it from the negatives,
leading to more stable learning and higher-quality trajectories.
Therefore, we adopt the latter design choice to define our InfoNCE objective as:
\begin{equation}
\begin{aligned}
\mathcal{L}_{\text{c}} &= -\log \frac{\exp(\ell^{+})}{\exp(\ell^{+})+\sum_{j=1}^{K}\exp(\ell^{-}_{j})}, \\
\ell^{+} &=-\mathcal{D}(q^{+},p^{c})/\tau, \\
\ell^{-}_{j} &=-\mathcal{D}(q^{+},q^{-}_{j})/\tau, 
\end{aligned}
\end{equation}
where $\mathcal{D}(\cdot,\cdot)$ is the Wasserstein-2 distance between Gaussians and $\tau$ is a temperature.

\begin{table*}[t]
\centering
\caption{Comparison with SOTA approaches on the closed-loop Bench2Drive benchmark on CARLA simulator. $\uparrow$ means the higher the better. DS is taken as the primary metric in comparison and we rank all the methods accordingly, with bold indicating best performance.}
\resizebox{\textwidth}{!}{
\begin{tabular}{lr|c|>{\columncolor{pink!20}}cccc}
\toprule
\textbf{Method} & \textbf{Venue} & \textbf{Scheme} & \textbf{DS$\uparrow$} & \textbf{SR(\%)$\uparrow$} & \textbf{Efficiency$\uparrow$} & \textbf{Comfortness$\uparrow$} \\
\midrule
VAD~\cite{Vad} & ICCV 2023 & IL & 42.35 & 15.00 & 157.94 & 46.01 \\
SparseDrive~\cite{Sparsedrive} & ICRA 2025 & IL & 44.54 & 16.71 & 170.21 & 48.63 \\
GenAD~\cite{Genad} & ECCV 2024 & IL & 44.81 & 15.90 & - & - \\
UniAD~\cite{UniAD} & CVPR 2023 & IL & 45.81 & 16.36 & 129.21 & 43.58 \\
MomAD~\cite{Momad} & CVPR 2025 & IL & 47.91 & 18.11 & 174.91 & \underline{51.20} \\
UAD~\cite{UAD} & T-PAMI 2025 & IL & 49.22 & 20.45 & 189.53 & \textbf{52.71} \\
BridgeAD~\cite{BridgeAD} & CVPR 2025 & IL & 50.06 & 22.73 & - & - \\
TCP~\cite{TCP} & NeurIPS 2022 & IL & 59.90 & 30.00 & 76.54 & 18.08 \\
WoTE~\cite{WoTE} & ICCV 2025 & IL & 61.71 & 31.36 & - & - \\
DriveDPO~\cite{Drivedpo} & NeurIPS 2025 & IL \& RL & 62.02 & 30.62 & 166.80 & 26.79 \\
ThinkTwice~\cite{ThinkTwice} & CVPR 2023 & IL & 62.44 & 31.23 & 69.33 & 16.22 \\
DriveTransformer~\cite{Drivetransformer} & ICLR 2025 & IL & 63.46 & 35.01 & 100.64 & 20.78 \\
DriveAdapter~\cite{Driveadapter} & ICCV 2023 & IL & 64.22 & 33.08 & 70.22 & 16.01 \\
Raw2Drive~\cite{Raw2Drive} & NeurIPS 2025 & RL & 71.36 & 50.24 & \underline{214.17} & 22.42 \\
Hydra-NeXt~\cite{HydraNeXt} & ICCV 2025 & IL & 73.86 & 50.00 & 197.76 & 20.68 \\
HiP-AD~\cite{Hip_ad} & ICCV 2025 & IL & 86.77 & 69.09 & 203.12 & 19.36 \\
\textbf{RaWMPC w/o Warm-up} & - & PC & \underline{87.34} & \underline{69.62} & 203.25 & 30.95 \\
\textbf{RaWMPC} & - & PC & \textbf{88.31} & \textbf{70.48} & 206.85 & 32.65 \\
\midrule
\textbf{Pretrained VLM-based Approach} \\
ReAL-AD~\cite{ReAL-AD} & ICCV 2025 & IL & 41.17 & 11.36 & - & - \\
Dual-AEB~\cite{Dual-AEB} & ICRA 2025 & IL & 45.23 & 10.00 & - & - \\
ETA~\cite{Eta} & ICCV 2025 & IL & 74.33 & 48.33 & 186.04 & 25.77 \\
VLR-Drive~\cite{VLR-Drive} & ICCV 2025 & IL & 75.01 & 50.00 & 122.52 & 0.59 \\
ORION~\cite{Orion} & ICCV 2025 & IL & 77.74 & 54.62 & 151.48 & 17.38 \\
SimLingo~\cite{Simlingo} & CVPR 2025 & IL & 85.94 & 66.82 & \textbf{244.18} & 30.76 \\
\bottomrule
\end{tabular}}
\label{tab:bench2drive_closedloop}
\end{table*}

\subsubsection{Overall Loss of Action Proposal Network}
The total loss of our cVAE combines reconstruction, KL regularization, and contrastive loss:
\begin{equation}
\begin{aligned}
\mathcal{L}_{\text{total}} &= \mathbb{E}_{z\sim q^{+}}\!\left[-\log p_\psi(\mathbf{A}^{+}\mid z,\mathbf{s})\right] \\
& + \beta\, D_{\text{KL}}\!\left(q^{+}\,\|\,p^{c}\right)
+\lambda\,\mathcal{L}_{\text{c}} .
\end{aligned}
\end{equation}
This self-evaluation distillation trains a fast proposal policy that generates candidate action sequences consistent with RaWMPC's evaluations, eliminating the need for expert demonstrations during policy learning.

\section{Experiments}
In this section, we present a comprehensive performance comparison between the proposed framework and state-of-the-art methods. We also conduct extensive ablation studies to assess the effectiveness of our predictive control approach.

\subsection{Benchmarks}

Following prior works~\cite{WoTE,Drivedpo,HydraNeXt}, we evaluate RaWMPC on two widely used benchmarks: \textbf{Bench2Drive}~\cite{Bench2drive} and \textbf{NAVSIM}~\cite{Navsim}.
They are complementary: Bench2Drive provides fully interactive closed-loop evaluation in CARLA~\cite{Carla} with dense annotations, while NAVSIM evaluates large-scale real-world planning via a data-driven, non-reactive simulation-based short-horizon rollout with safety- and progress-aware metrics.

\noindent
\textbf{Bench2Drive.}
Bench2Drive is a CARLA Leadboard v2 closed-loop benchmark for multi-ability stress testing under complex interactions (e.g., cut-ins, overtakes, detours, emergency braking, and give-way).
Its official dataset contains $\sim$2M fully annotated frames from short clips spanning 44 scenarios, 23 weather conditions, and 12 towns; the commonly used \textit{Base} training set contains 1K clips.
Closed-loop evaluation is performed on 220 short routes (each focused on a single scenario), enabling stable and fine-grained comparison.
We report four official metrics: \textit{Driving Score (DS)}, \textit{Success Rate (SR)}, \textit{Efficiency}, and \textit{Comfortness}.
\textit{DS} is the primary aggregate score with penalties for safety and rule violations; \textit{SR} measures successful completion; \textit{Efficiency} reflects progress; and \textit{Comfortness} captures motion smoothness.

\begin{table*}[t]
\centering
\caption{Comparison with the SOTA approaches on NAVSIM test set. $\uparrow$ means the higher the better.
PDMS is taken as the primary metric in comparison and we rank all the methods accordingly,
with bold indicating best performance.}
\resizebox{\textwidth}{!}{
\begin{tabular}{lr|c|ccccc|>{\columncolor{pink!20}}c}
\toprule
\textbf{Method} & \textbf{Venue} & \textbf{Scheme} & \textbf{NC$\uparrow$} & \textbf{DAC$\uparrow$} & \textbf{EP$\uparrow$} & \textbf{TTC$\uparrow$} & \textbf{C$\uparrow$} & \textbf{PDMS$\uparrow$} \\
\midrule
Human & - & - & 100 & 100 & 87.5 & 100 & 99.9 & 94.8 \\
\midrule
DrivingGPT~\cite{DrivingGPT} & ICCV 2025 & IL & \underline{98.9} & 90.7 & 79.7 & 94.9 & 100.0 & 82.4 \\
UniAD~\cite{UniAD} & CVPR 2023 & IL & 97.8 & 91.9 & 78.8 & 92.9 & 100.0 & 83.4 \\
Latent TransFuser~\cite{Transfuser} & T-PAMI 2023 & IL & 97.4 & 92.8 & 79.0 & 92.4 & 100.0 & 83.8 \\
PARA-Drive~\cite{PARA_Drive} & CVPR 2024 & IL & 97.9 & 92.4 & 79.3 & 93.0 & 99.8 & 84.0 \\
TransFuser~\cite{Transfuser} & T-PAMI 2023 & IL & 97.7 & 92.7 & 79.8 & 92.7 & 100.0 & 84.5 \\
LAW~\cite{LAW} & ICLR 2025 & IL & 96.4 & 95.4 & 81.7 & 88.7 & 99.9 & 84.6 \\
World4Drive~\cite{World4Drive} & ICCV 2025 & IL & 97.4 & 94.3 & 79.9 & 92.8 & 100.0 & 85.1 \\
DiffusionDrive~\cite{Diffusiondrive} & CVPR 2025 & IL & 98.2 & 96.2 & \textbf{88.2} & 94.7 & 100.0 & 88.1 \\
WoTE~\cite{WoTE} & ICCV 2025 & IL & 98.5 & 96.8 & 81.9 & 94.9 & 99.9 & 88.3 \\
Hydra-NeXt~\cite{HydraNeXt} & ICCV 2025 & IL & 98.1 & 97.7 & 81.8 & 94.6 & 100.0 & 88.6 \\
UAD~\cite{UAD} & T-PAMI 2025 & IL & \textbf{99.5} & 96.9 & 78.8 & \textbf{97.5} & 100.0 & 89.3 \\
DriveDPO~\cite{Drivedpo} & NeurIPS 2025 & IL \& RL & 98.5 & 98.1 & 84.3 & 94.8 & 99.9 & 90.0 \\
GoalFlow~\cite{Goalflow} & CVPR 2025 & IL & 98.4 & \textbf{98.3} & 85.0 & 94.6 & 100.0 & 90.3 \\
\textbf{RaWMPC w/o Warm-up} & - & PC & 98.3 & \underline{98.2} & 85.3 & 94.5 & 99.9 & \underline{90.5} \\
\textbf{RaWMPC} & - & PC & \underline{98.9} & \textbf{98.3} & \underline{86.1} & \underline{95.6} & 99.9 & \textbf{91.3} \\
\bottomrule
\end{tabular}}
\label{tab:NAVSIM}
\end{table*}

\noindent
\textbf{NAVSIM.}
NAVSIM benchmarks sensor-based planning on large-scale real-world data built on OpenScene (a planning-oriented reprocessing of nuPlan logs).
The task predicts a 4-second future ego trajectory (typically 8 waypoints) given a short history (e.g., 1.5 seconds) of observations.
Following prior works~\cite{Drivedpo,WoTE}, we use the official splits: Navtrain ($\sim$103K samples) and Navtest ($\sim$12K samples).
We report NAVSIM metrics including \textit{NC}, \textit{DAC}, \textit{EP}, \textit{TTC}, \textit{C}, and the primary score \textit{PDMS}, where $\mathrm{PDMS}=\mathrm{NC}\cdot\mathrm{DAC}\cdot(5\mathrm{EP}+5\mathrm{TTC}+2\mathrm{C})/12$.
These metrics jointly capture safety, compliance, progress, and motion quality.
All results are computed with the official toolkits and recommended splits.

Our experiments combine offline warm-up with online simulator interaction, leveraging RaWMPC to learn predictive dynamics and risk-aware decision making. We adopt Bench2Drive for interactive closed-loop evaluation and NAVSIM for large-scale real-world generalization.
We further conduct ablations to analyze key training components and strategies.

\subsection{Implementation Details}

\textbf{Network architecture.} We employ a pretrained ViT~\cite{Dinov3} as the vision encoder and use the SegViT segmentation head~\cite{SegViT} as the segmentation decoder. BEV features are extracted from multi-view images using the query-based view transformer~\cite{Bevformer}. Following previous works~\cite{WoTE,Transfuser}, the input image resolution is $1024\times256$, and the BEV feature map resolution is $256\times256$. We use down-sampling factors of 32 (front-view) and 16 (BEV), resulting in 512 visual tokens $\mathbf{i}_{t}$ (256 per branch). Measurement inputs are encoded through an MLP-based encoder into 4 measurement tokens $\mathbf{m}_{t}$, and driving actions, represented as three scalar values (steer, throttle, brake), are transformed into 3 action tokens $\mathbf{a}_{t}$ via a linear layer. We implement the world model as a transformer with 4 layers and 8 attention heads. RaWMPC predicts $H{=}10$ future steps conditioned on the past 5 observed steps, and evaluates $N{=}10$ candidate action sequences proposed by the distilled action proposal network (described in Section~\ref{sec:guidance}) using the cost in Eq.~\eqref{eq_cost} during inference. During action selection, we use $\eta_k=\max(2^{-k+1},1/8)$ to downweight distant predictions, and set the event severity weights to $\lambda_j={10,15,30}$ for off-lane driving, traffic-sign violations, and collisions, respectively.

\noindent
\textbf{Training.} RaWMPC is trained with the two-stage risk-aware interactive training strategy described in Section~\ref{sec:RaIT}. We first perform an offline warm-up using 10\% of the training data (100 clips for Bench2Drive and 10K samples for NAVSIM), and then refine the model via online interaction using the proposed risk-aware training scheme. The random-sampling probability $\varepsilon_1$ is linearly annealed from 1 to 0, while the risk-sampling probability $\varepsilon_2$ is linearly increased from 0 to 0.3. We maintain a replay buffer of size 10K frames to store recent interaction data (e.g., RGB images, semantic segmentation, ego measurements, and traffic-event annotations).
We use segment-wise sampling of horizon-$H{=}10$ action sequences to ensure temporal continuity. In good/bad modes, we rank the $N_s{=}50$ candidates by cost and sample from the bottom/top-$K{=}5$ sets using temperatures $\tau_g=0.5$ and $\tau_b=1.0$.
An episode terminates when any of the following conditions is met: (1) the ego vehicle incurs 3 collisions, (2) the ego vehicle goes off-road or remains stuck for 100 consecutive steps, or (3) the ego vehicle successfully completes the route. Across offline warm-up and online interaction, we train RaWMPC using a total of 1K clips on Bench2Drive and 100K samples on NAVSIM (comparable in scale to the official training sets for fair comparison), on four NVIDIA A100 GPUs. We use Adam~\cite{Adam} with an initial learning rate of $10^{-4}$ decayed to $10^{-5}$, and a batch size of 16.

\noindent
\textbf{Self-evaluation distillation.} For self-evaluation distillation (Section~\ref{sec:guidance}), the action proposal network is implemented as a cVAE~\cite{cVAE} with a 32-dimensional latent space. During contrastive training, we sample $N_s{=}50$ action sequences, treat the minimum-cost sequence as the positive example $\mathbf{A}^+$, and use the top-$K{=}5$ highest-cost sequences as negatives. We optimize the proposal network with the objective in Section~\ref{sec:guidance}, using temperature $\tau{=}0.3$, a KL weight schedule $\beta:0\rightarrow0.1$, and a contrastive weight $\lambda{=}0.1$. At inference, we sample $N{=}10$ candidates from the cVAE decoder and select the final action by minimizing the predicted cost (Eq.~\eqref{eq_cost}) under the world-model rollout.

\subsection{Comparison with the State-of-the-Art}

\begin{figure*}
    \centering
    \includegraphics[width=\linewidth]{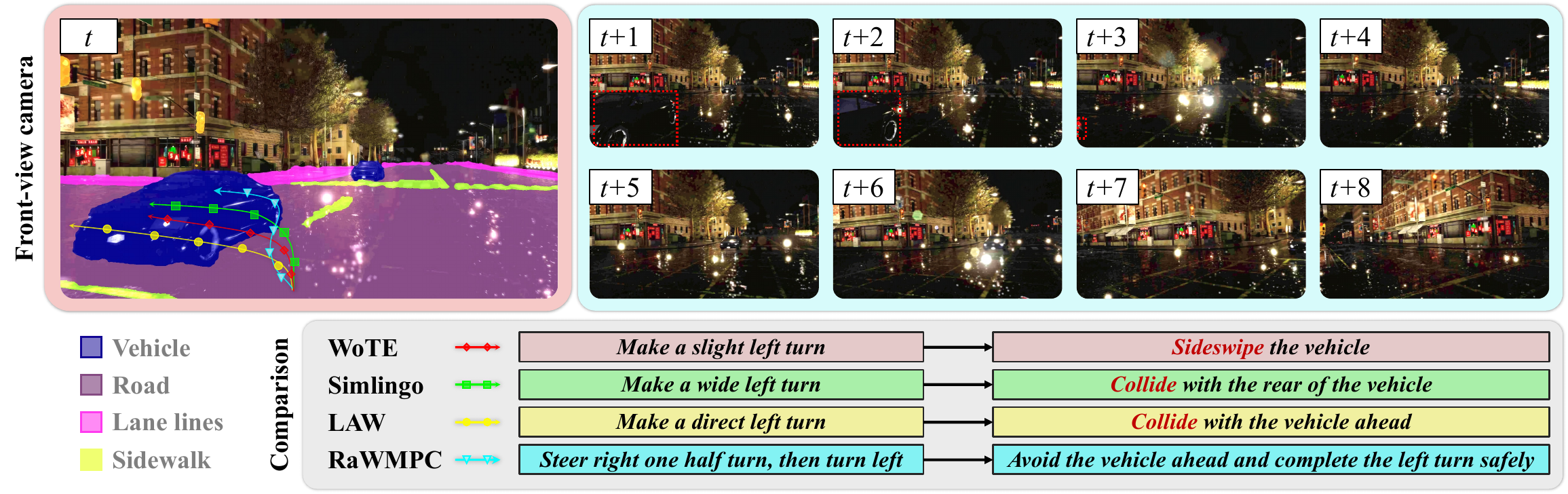}
    \caption{
    \textbf{Qualitative comparison under weather-induced domain shift (\textit{Sunny-only} $\rightarrow$ \textit{Rainy}).}
All methods are trained on \textit{Sunny-only} data and evaluated in \textit{Rainy} conditions.
LAW \cite{LAW} misses the lead vehicle, causing a severe frontal collision.
WoTE \cite{WoTE} and SimLingo \cite{Simlingo} reduce severity by evasive maneuvers but still collide due to degraded perception–decision reliability and weak safety-margin enforcement.
RaWMPC avoids collisions by selecting the minimum-risk predictive-control action under uncertainty.
    }
    \label{fig:vis_ood}
\end{figure*}

\begin{table*}[t]
    \centering
    \setlength{\tabcolsep}{5.0pt}
    \caption{Performance comparison under domain shift.
    All methods are trained on either \textit{Sunny only} or \textit{Sunny \& Rainy} data, and evaluated exclusively on \textit{Rainy} scenarios.
    $\uparrow$ indicates higher is better.}
    \begin{tabular}{lrcclcc}
        \toprule
        \multirow{2.5}{*}{\textbf{Method}} 
        & \multirow{2.5}{*}{\textbf{Venue}} 
        & \multirow{2.5}{*}{\textbf{Scheme}} 
        & \multirow{2.5}{*}{\textbf{Training Data}} 
        & \multicolumn{2}{c}{\textbf{Tested on Rainy}} \\
        \cmidrule(lr){5-6}
        & & & & \textbf{DS$\uparrow$} & \textbf{SR(\%)$\uparrow$} \\
        \midrule
        \multirow{2}{*}{LAW} 
        & \multirow{2}{*}{ICLR 2025} 
        & \multirow{2}{*}{IL} 
        & Sunny \& Rainy 
        & 34.54 & 7.09 \\
        & & & Sunny only 
        & 23.58 & 3.56 \\
        \midrule
        \multirow{2}{*}{WoTE} 
        & \multirow{2}{*}{ICCV 2025} 
        & \multirow{2}{*}{IL} 
        & Sunny \& Rainy 
        & 36.54 & 7.85 \\
        & & & Sunny only 
        & 28.65 & 5.21 \\
        \midrule
        \multirow{2}{*}{SimLingo (Pretrained-VLM)} 
        & \multirow{2}{*}{CVPR 2025} 
        & \multirow{2}{*}{IL} 
        & Sunny \& Rainy 
        & 51.69 & 13.68 \\
        & & & Sunny only 
        & 33.49 & 8.97 \\
        \midrule
        \multirow{2}{*}{\textbf{RaWMPC}} 
        & \multirow{2}{*}{--} 
        & \multirow{2}{*}{PC} 
        & Sunny \& Rainy 
        & \textbf{53.67} & \textbf{14.96} \\
        & & & Sunny only 
        & \textbf{41.36} & \textbf{10.83} \\
        \bottomrule
    \end{tabular}
    \label{tab:domain_shift_rainy}
\end{table*}

In this section,
we compare RaWMPC with state-of-the-art end-to-end driving methods on the closed-loop Bench2Drive benchmark in CARLA and the NAVSIM test set.
To reflect our main claim, we report two training settings:
(i) \textbf{w/o warm-up}, where RaWMPC is trained without using any offline logged video,
and (ii) the \textbf{full} setting,
where a small set of logged driving trajectories is used as an optional warm start.
The warm start empirically accelerates convergence and further improves final performance, while RaWMPC already surpasses prior state-of-the-art even without it.

\subsubsection{Evaluation on Bench2Drive}
Table~\ref{tab:bench2drive_closedloop} illustrates the closed-loop results on Bench2Drive.
RaWMPC achieves the best overall performance among all compared methods, reaching \textbf{88.31} DS and \textbf{70.48}\% SR in the full setting.
More importantly, even \textbf{without warm-up} (i.e., without logged trajectories), RaWMPC still attains \underline{87.34} DS and 69.62\% SR, surpassing strong recent baselines such as HiP-AD~\cite{Hip_ad} (86.77 DS / 69.09\% SR) and the pretrained-VLM method SimLingo~\cite{Simlingo} (85.94 DS / 66.82\% SR).
In addition, RaWMPC maintains competitive efficiency and achieves higher comfortness than most high-performing closed-loop agents, indicating that the gains are not obtained by aggressive maneuvers but by more reliable decision-making.

\subsubsection{Evaluation on NAVSIM}
Table~\ref{tab:NAVSIM} summarizes the results on NAVSIM.
RaWMPC achieves the highest \textbf{PDMS of 91.3} among all learning-based methods.
Without warm-up, RaWMPC still reaches \underline{90.5} PDMS, already outperforming previous best methods (e.g., GoalFlow~\cite{Goalflow}: 90.3).
The warm start further improves PDMS (90.5$\rightarrow$91.3), consistent with the observation that a small amount of logged trajectories can accelerate convergence and improve performance, while not being required to achieve state-of-the-art results.

\subsubsection{Generalization under weather-induced domain shift}
To evaluate robustness beyond the training distribution, we conduct a weather-shift study where all methods are evaluated exclusively on \textit{Rainy} scenarios while being trained on either \textit{Sunny only} or \textit{Sunny \& Rainy} data (Table~\ref{tab:domain_shift_rainy}).
A key observation is that imitation-based methods are sensitive to training-domain coverage: when rainy conditions are absent from training (Sunny-only), their performance drops notably, reflecting limited transfer to previously unseen environments.
In contrast, RaWMPC achieves the best DS and SR under both training regimes, and notably still significantly outperforms strong IL baselines (LAW, WoTE) and SimLingo when trained on \textit{Sunny only} (\textit{i.e.}, facing an unseen rainy target domain).
Moreover, compared with SimLingo, RaWMPC exhibits substantially smaller degradation when rainy data is removed from training,
indicating stronger robustness to previously unseen conditions.


Figure~\ref{fig:vis_ood} shows a qualitative example of this \textit{Sunny-only} $\rightarrow$ \textit{Rainy} shift.
LAW fails to recognize the lead vehicle under the altered visual conditions and results in a high-severity frontal collision.
WoTE and SimLingo attempt evasive maneuvers that reduce the impact \emph{relative to a direct frontal crash}, yet the scene still ends in a side-swipe/rear collision.
A plausible explanation is that the rainy shift degrades the reliability of the perception--decision stack, while the downstream policy does not explicitly optimize for minimum-risk clearance under uncertainty, yielding insufficient safety margins during close-proximity avoidance (e.g., inaccurate motion anticipation or mismatched ego response on wet roads).
By contrast, RaWMPC explicitly evaluates the predicted consequences of candidate action sequences with a risk-aware world model and selects the minimum-risk predictive-control behavior, maintaining safe clearance under uncertainty.

We attribute this advantage to RaWMPC's risk-aware predictive-control formulation: instead of reproducing expert actions, RaWMPC learns \emph{risk-awareness from interaction} and selects actions by minimizing predicted risk via the learned world model.
Such an objective encourages transferable decision principles (e.g., maintaining safe margins and acting conservatively under uncertainty) that remain effective when appearance and dynamics change across domains.
Thus, RaWMPC is less dependent on exhaustive expert action coverage for corner cases, which better matches the long-tail nature of real-world deployment where unseen scenarios are inevitable.

\begin{figure*}[!t]
    \centering
    \includegraphics[width=\linewidth]{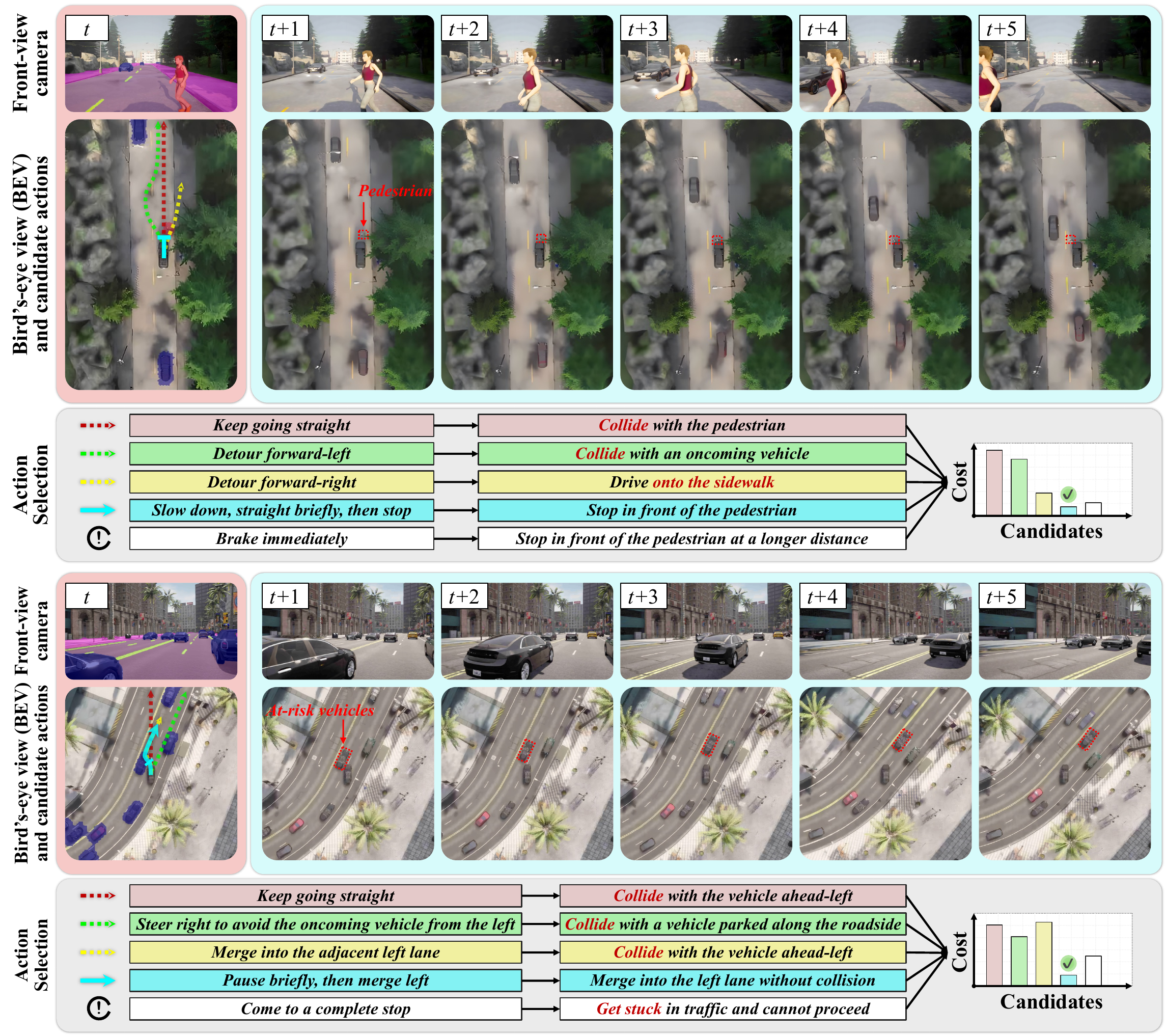}
    \caption{\textbf{Visualization of the predictive control procedure}.
At time $t$, we show the front-view and the BEV images (with segmentation).
Dashed curves indicate candidate actions and highlighted agents denote key risks.
Rollouts from $t+1$ to $t+5$ illustrate predicted consequences,
and the bottom panel reports each action’s outcomes and costs (\textit{e.g.}, collision, sidewalk intrusion, stopping distance).
\textbf{\textit{Scenario 1}}: RaWMPC slows down, proceeds briefly, then stops for the pedestrian.
\textbf{\textit{Scenario 2}}: RaWMPC waits briefly, then turns left to avoid both the front-left vehicle and the parked car.
    }
    \label{fig:vis1}
\end{figure*}

\subsubsection{Qualitative visualization of predictive control}
We provide some visualization results of the predictive control procedure of RaWMPC in Fig.~\ref{fig:vis1}.
Given RGB observations and high-level navigation commands (e.g., keep going straight or merge left), our generative policy proposes a small set of candidate action sequences (e.g., keep going straight, detour, brake, and lane change).
For each candidate, the risk-aware world model predicts the near-future semantic traffic state and the decoding module evaluates its consequence from both task progress and safety perspectives, including collision risk, off-lane/sidewalk intrusion risk, and progress-related penalties such as getting stuck in traffic.
The final action is selected by comparing these predicted consequences and choosing the minimum-cost one under the navigation goal.



In the \textbf{\textit{first}} case, going straight collides with a crossing pedestrian, while detours either collide with an oncoming vehicle or drive onto the sidewalk; RaWMPC chooses \emph{slow down, straight briefly, then stop} to safely stop in front of the pedestrian (instead of an overly conservative early brake).
In the \textbf{\textit{second}} case, going straight or merging left immediately causes a collision, steering right hits a parked vehicle, and stopping leads to a deadlock; thus RaWMPC selects \emph{pause briefly, then merge left} for a collision-free merge. These cases demonstrate that RaWMPC can proactively avoid risky behaviors by explicitly forecasting and comparing action consequences, rather than merely following a single command or relying on a fixed fallback maneuver.


\subsection{Ablation Study}

In this section, we provide comprehensive ablation studies of the proposed approach using the Bench2Drive dataset.

\subsubsection{Analysis of Framework}

\begin{table}[t]
    \centering
    \setlength{\tabcolsep}{6pt}
    \caption{Ablation study on the proposed framework.}
    \resizebox{\columnwidth}{!}{%
    \begin{tabular}{lll}
        \toprule
        \multirow{2.5}{*}{\textbf{Method}} & \multicolumn{2}{c}{\textbf{Metrics}} \\
        \cmidrule(lr){2-3}
        & \textbf{DS$\uparrow$} & \textbf{SR(\%)$\uparrow$} \\
        \midrule
        Entire RaWMPC (\textbf{Ours}) & \textbf{88.31} & \textbf{70.48} \\
        w/o Semantic Guidance
            & 82.36 {\tiny\color{red} -5.95}
            & 62.69 {\tiny\color{red} -7.79} \\
        w/o Segmentation Decoder
            & 70.85 {\tiny\color{red} -17.46}
            & 48.95 {\tiny\color{red} -21.53} \\
        w/o Action Selection
            & 61.35 {\tiny\color{red} -26.96}
            & 30.98 {\tiny\color{red} -39.50} \\
        \bottomrule
    \end{tabular}}%
    \label{tab:ablation_framework}
\end{table}

Table~\ref{tab:ablation_framework} analyzes core components aligned with our model design (Sec.~\ref{sec:pipeline}).
\emph{w/o Semantic Guidance} removes semantic-guided event decoding,
\textit{i.e.}, the fusion of semantic attention from the segmentation decoder into the event decoder.
This leads to a clear drop (DS 88.31$\rightarrow$82.36, SR 70.48\%$\rightarrow$62.69\%), showing that accurate safety-event prediction is crucial for risk-aware cost evaluation.
\emph{w/o Segmentation Decoder} further removes the segmentation decoding branch and its supervision, resulting in a large degradation (DS 70.85 / SR 48.95\%), which indicates that forecasting high-level semantics is essential for reliable long-horizon rollouts.
\emph{w/o Action Selection} disables predictive control in Eq. \eqref{eq:action_selection} (bypassing cost-based ranking in Eq. \eqref{eq_cost}) and directly executes the proposal/guidance output, causing the most severe collapse (DS 61.35 / SR 30.98\%).
This confirms that selecting actions by explicitly evaluating predicted long-horizon consequences is the key to RaWMPC.

\subsubsection{Analysis of Risk-Aware Training}

Table~\ref{tab:ablation_risk_aware} evaluates the risk-aware interaction training strategy used to refine the world model.
Our \emph{risk-aware sampling} follows the design in Sec. \ref{sec:RaIT}: besides random exploration, it uses the current RaWMPC to score candidate action sequences and deliberately collects both \textbf{good} (low-cost) and \textbf{bad} (high-cost) rollouts, improving coverage of rare safety-critical outcomes.
Replacing it with \emph{$\epsilon$-greedy sampling} (random with probability $\epsilon_1$, otherwise only selecting low-cost rollouts) reduces performance (DS 83.86 / SR 61.74\%), showing that excluding high-cost failures weakens learning of risky consequences.
Pure \emph{random sampling} further degrades results (DS 70.41 / SR 46.82\%), indicating that unguided data collection is substantially less efficient for learning long-horizon consequences.

\begin{table}[t]
    \centering
    \setlength{\tabcolsep}{6pt}
    \caption{Ablation study on the risk-aware training.}
    \resizebox{\columnwidth}{!}{%
    \begin{tabular}{lll}
        \toprule
        \multirow{2.5}{*}{\textbf{Method}} & \multicolumn{2}{c}{\textbf{Metrics}} \\
        \cmidrule(lr){2-3}
        & \textbf{DS$\uparrow$} & \textbf{SR(\%)$\uparrow$} \\
        \midrule
        Risk-aware Sampling (\textbf{Ours}) & \textbf{88.31} & \textbf{70.48} \\
        $\epsilon$-Greedy Sampling
            & 83.86 {\tiny\color{red} -4.45}
            & 61.74 {\tiny\color{red} -8.74} \\
        Random Sampling
            & 70.41 {\tiny\color{red} -17.90}
            & 46.82 {\tiny\color{red} -23.66} \\
        \bottomrule
    \end{tabular}}%
    \label{tab:ablation_risk_aware}
\end{table}
          
\begin{table}[t]
    \centering
    \setlength{\tabcolsep}{6pt}
    \caption{Results obtained using different action supervision in policy learning.}
    \resizebox{\columnwidth}{!}{%
    \begin{tabular}{lll}
        \toprule
        \multirow{2.5}{*}{\textbf{Policy Learning Data}} & \multicolumn{2}{c}{\textbf{Metrics}} \\
        \cmidrule(lr){2-3}
        & \textbf{DS$\uparrow$} & \textbf{SR(\%)$\uparrow$} \\
        \midrule
        Pos. \& Neg. Actions (\textbf{Ours}) & \textbf{88.31} & \textbf{70.48} \\
        Expert Actions
            & 86.75 {\tiny\color{red} -1.56}
            & 68.25 {\tiny\color{red} -2.23} \\
        Only Positive Actions
            & 83.65 {\tiny\color{red} -4.66}
            & 66.52 {\tiny\color{red} -3.96} \\
        \bottomrule
    \end{tabular}}%
    \label{tab:ablation_policy_learning}
\end{table}

\subsubsection{Analysis of Self-Evaluation Distillation}
Table~\ref{tab:ablation_policy_learning} studies how to train the action proposal network in Sec. \ref{sec:guidance}.
Our default setting uses RaWMPC as a self-evaluator to pseudo-label actions:
the lowest-cost sequence is treated as a positive, while high-cost sequences serve as negatives,
which yields the best performance (DS 88.31 / SR 70.48\%).
Training the proposal network only with \emph{expert actions} slightly degrades performance (DS 86.75 / SR 68.25\%),
suggesting that self-evaluated targets align better with the predictive-control objective than direct imitation targets.
Using \emph{only positive actions} further drops performance (DS 83.65 / SR 66.52\%),
indicating that explicitly contrasting against high-risk negatives is important for preventing unsafe candidates and improving downstream selection.

\subsubsection{Discussion on Prediction Horizon}

Table~\ref{tab:prediction_horizon} ablates the planning horizon $H$ used in the world-model rollout and cost evaluation (Eq.~\eqref{eq_cost}).
Short horizons fail to capture delayed consequences,
leading to poor performance (H=1: DS 57.85 / SR 28.64\%; H=5: DS 74.98 / SR 49.52\%).
Increasing to H=10 yields the best results (DS 88.31 / SR 70.48\%), as it provides sufficient look-ahead for risk assessment while keeping prediction uncertainty manageable.
Further increasing to H=15 degrades performance (DS 82.34 / SR 62.38\%), likely due to accumulated rollout errors that affect cost-based ranking.

\begin{table}[t]
    \centering
    \setlength{\tabcolsep}{6pt}
    \caption{Results obtained using different amounts of prediction horizon.}
    \begin{tabular}{lll}
        \toprule
        \multirow{2.5}{*}{\textbf{Prediction Horizon}} & \multicolumn{2}{c}{\textbf{Metrics}} \\
        \cmidrule(lr){2-3}
        & \textbf{DS$\uparrow$} & \textbf{SR(\%)$\uparrow$} \\
        \midrule
        H=1
            & 57.85 {\tiny\color{red} -30.46}
            & 28.64 {\tiny\color{red} -41.84} \\
        H=5
            & 74.98 {\tiny\color{red} -13.33}
            & 49.52 {\tiny\color{red} -20.96} \\
        H=10 (\textbf{Ours})
            & \textbf{88.31}
            & \textbf{70.48} \\
        H=15
            & 82.34 {\tiny\color{red} -5.97}
            & 62.38 {\tiny\color{red} -8.10} \\
        \bottomrule
    \end{tabular}
    \label{tab:prediction_horizon}
\end{table}

\subsubsection{Discussion on Warm-up}

\begin{table}[t]
    \centering
    \setlength{\tabcolsep}{6pt}
    \caption{Results obtained using different amounts of offline learning data in warm-up.}
    \begin{tabular}{lll}
        \toprule
        \multirow{2.5}{*}{\textbf{Warm-up Data}} & \multicolumn{2}{c}{\textbf{Metrics}} \\
        \cmidrule(lr){2-3}
        & \textbf{DS$\uparrow$} & \textbf{SR(\%)$\uparrow$} \\
        \midrule
        0\%
            & 87.34 {\tiny\color{red} -0.97}
            & 69.62 {\tiny\color{red} -0.86} \\
        10\%
            & \textbf{88.31}
            & \textbf{70.48} \\
        20\% 
            & 88.09 {\tiny\color{red} -0.22}
            & 70.32 {\tiny\color{red} -0.16} \\
        30\%
            & 86.95 {\tiny\color{red} -1.36}
            & 68.52 {\tiny\color{red} -1.96} \\
        \bottomrule
    \end{tabular}
    \label{tab:warmup}
\end{table}

Table~\ref{tab:warmup} ablates the fraction of offline logged trajectories used for warm-up before interactive training. In this study, we keep the total number of training samples fixed and vary only the proportion allocated to offline warm-up data.
Without warm-up (0\%), performance drops (DS 87.34 / SR 69.62\%), suggesting that training from scratch leads to less reliable rollouts and a less stable early optimization stage. Using a small amount of logged trajectories (10\%) yields the best results (DS 88.31 / SR 70.48\%), indicating that a light warm-up provides useful predictive priors (e.g., basic dynamics modeling and perception decoding) that improve long-horizon rollout quality and downstream control. However, further increasing the warm-up ratio begins to degrade performance (20\%: DS 88.09 / SR 70.32\%; 30\%: DS 86.95 / SR 68.52\%). We attribute this trend to the fact that offline logged trajectories are strongly biased toward safe, human-like behaviors and contain few hazardous events. Consequently, allocating too much data to offline warm-up reduces the opportunities for subsequent online interaction to explore unconventional actions and collect safety-critical failures—especially in dangerous scenarios—which are essential for learning robust risk awareness.

\subsubsection{Discussion on Control Pattern} 

Table~\ref{tab:ablation_control_pattern} compares predictive control to directly optimizing a policy with model-based reinforcement learning (RL).
Predictive control achieves substantially higher performance (DS 88.31 / SR 70.48\%) than model-based RL (DS 73.58 / SR 51.85\%).
This validates the benefit of evaluating candidate action sequences via decoded future outcomes (segmentation, events, ego-states) and selecting the minimum-cost one,
rather than relying on end-to-end policy optimization alone.

\begin{table}[t]
    \centering
    \setlength{\tabcolsep}{6pt}
    \caption{Ablation study on the control pattern with learned world model.}
    \resizebox{\columnwidth}{!}{%
    \begin{tabular}{lll}
        \toprule
        \multirow{2.5}{*}{\textbf{Method}} & \multicolumn{2}{c}{\textbf{Metrics}} \\
        \cmidrule(lr){2-3}
        & \textbf{DS$\uparrow$} & \textbf{SR(\%)$\uparrow$} \\
        \midrule
        Predictive Control (\textbf{Ours})
            & \textbf{88.31}
            & \textbf{70.48} \\
        Reinforcement Learning
            & 73.58 {\tiny\color{red} -14.73}
            & 51.85 {\tiny\color{red} -18.63} \\
        \bottomrule
    \end{tabular}}%
    \label{tab:ablation_control_pattern}
\end{table}

\subsubsection{World-Model Prediction Accuracy}
Table \ref{tab:prediction_acc} reports the event prediction quality used by the cost function in Eq.~\eqref{eq_cost}.
The predictor achieves high accuracy across event types (0.91--0.96) and strong recall on collision-related events (e.g., pedestrian collision recall 0.99), providing reliable signals for risk-aware evaluation.
We observe lower precision for some rare events (e.g., pedestrian collision precision 0.52), reflecting a conservative tendency with more false positives; in safety-critical driving, prioritizing recall can be preferable to missing hazards.

\begin{table}[t]
    \setlength\tabcolsep{3pt}
    \caption{Prediction accuracy of future traffic events with learned world model.}
	\centering
    \begin{tabular}{lcccc}
        \toprule
		& \multicolumn{3}{c}{\textbf{Collision}} & \multirow{2.5}{*}{\makecell{\textbf{Running}\\\textbf{Traffic Sign}}} \\
        \cmidrule(r){2-4}
		& \textbf{Pedestrian} & \textbf{Vehicle} & \textbf{Static} & \\
        \midrule
            Accuracy & 0.96 & 0.91 & 0.93 & 0.91 \\
            \midrule
            Recall & 0.99 & 0.84 & 0.89 & 0.84 \\ 
            Precision & 0.52 & 0.62 & 0.63 & 0.68 \\ 
        \bottomrule
	\end{tabular}
    \label{tab:prediction_acc}
\end{table}

\section{Conclusion}
In this work, we proposed \textbf{RaWMPC}, a risk-aware world-model predictive control framework for end-to-end autonomous driving that \emph{does not require expert action supervision}.
RaWMPC learns an action-conditioned world model to roll out multiple candidate behaviors, predicts future semantics and safety-critical events, and selects actions by explicitly minimizing a risk-aware cost.
To make rare-but-catastrophic outcomes predictable and avoidable, we introduced a \textbf{risk-aware interaction} strategy that intentionally collects both safe and hazardous rollouts, and we further proposed \textbf{self-evaluation distillation} to train an efficient action proposal policy using RaWMPC as a self-evaluator.
Extensive experiments on Bench2Drive and NAVSIM show that RaWMPC achieves state-of-the-art performance and stronger robustness under domain shift, even without offline warm-up, showing the potential to significantly reduce the reliance on costly real-world expert demonstrations.
For future work, we will explore domain adaptation and more efficient planning to better support real-world deployment and sim-to-real transfer.




\section*{Statements and Declarations}


\begin{itemize}
\item Competing interests.
The authors declare that they have no known competing financial interests or personal relationships that could have appeared to influence the work reported in this paper.
\item Data availability.
This work does not propose any new dataset. The datasets (Bench2Drive~\cite{Bench2drive} and NAVSIM~\cite{Navsim}) that support the findings of this study are openly available at the URLs: \href {https://github.com/Thinklab-SJTU/Bench2Drive/tree/main}{Bench2Drive} and \href {https://github.com/autonomousvision/navsim}{NAVSIM}.
\end{itemize}

\bibliographystyle{plainnat}
\bibliography{main}

\end{document}